\documentclass[conference]{IEEEtran}

\usepackage{cite}

\usepackage{soul}

\usepackage{amsmath}

\usepackage{algorithmic}

\usepackage{array}

\usepackage{listings}
\usepackage{algorithm} 
\usepackage{graphicx}

\usepackage{url}
\usepackage[flushleft]{threeparttable}

\ifCLASSOPTIONcompsoc
  \usepackage[caption=false,font=normalsize,labelfont=sf,textfont=sf]{subfig}
\else
  \usepackage[caption=false,font=footnotesize]{subfig}
\fi

\usepackage{hyperref}
\usepackage{url}

\usepackage{color}

\renewcommand{\baselinestretch}{0.99}

\begin{document}

\title{Caffeinated FPGAs: FPGA Framework For Convolutional Neural Networks}

\author{\IEEEauthorblockN{Roberto DiCecco\IEEEauthorrefmark{1},
Griffin Lacey\IEEEauthorrefmark{2},
Jasmina Vasiljevic\IEEEauthorrefmark{1}, 
Paul Chow\IEEEauthorrefmark{1},
Graham Taylor\IEEEauthorrefmark{2} and
Shawki Areibi\IEEEauthorrefmark{2}}
\IEEEauthorblockA{\IEEEauthorrefmark{1}University of Toronto, Department of Electrical and Computer Engineering, Ontario, Canada\\E-mail:\{dicecco1, vasiljev, pc\}@eecg.toronto.edu}
\IEEEauthorblockA{\IEEEauthorrefmark{2}University of Guelph, Ontario, Canada\\E-mail:\{laceyg, gwtaylor, sareibi\}@uoguelph.ca}}

\maketitle

\begin{abstract}
Convolutional Neural Networks (CNNs) have gained significant traction in the field of machine learning, particularly due to their high accuracy in visual recognition. Recent works have pushed the performance of GPU implementations of CNNs to significantly improve their classification and training times. With these improvements, many frameworks have become available for implementing CNNs on both CPUs and GPUs, with no support for FPGA implementations. In this work we present a modified version of the popular CNN framework Caffe, with FPGA support. This allows for classification using CNN models and specialized FPGA implementations with the flexibility of reprogramming the device when necessary, seamless memory transactions between host and device, simple-to-use test benches, and the ability to create pipelined layer implementations.  To validate the framework, we use the Xilinx SDAccel environment to implement an FPGA-based Winograd convolution engine and show that the FPGA  layer can be used alongside other layers running on a host processor to run several popular CNNs (AlexNet, GoogleNet, VGG A, Overfeat). The results show that our framework achieves 50 GFLOPS across 3 $\times$ 3 convolutions in the benchmarks. This is achieved within a practical framework, which will aid in future development of FPGA-based CNNs. 

\end{abstract}

\IEEEpeerreviewmaketitle

\section{Introduction}\label{section:intro}

Convolutional Neural Networks (CNNs) are highly accurate deep learning networks inspired by the mammalian visual cortex. A number of works explored
the implementation of CNNs on FPGAs~\cite{zhang, suda, qiu} to take advantage of their low-power, customizable and programmable fabric. While FPGA
implementations show promise in efficiently computing CNNs, they lack the infrastructure available for both CPUs and GPUs.  
This makes FPGAs inaccessible to deep learning scientists. There are many frameworks for CNN implementations, most of which provide 
support for CPU, GPU or the option of both~\cite{caffe}. These frameworks allow the programmer to launch any CNN model, 
and contain comprehensive tests for both layer-based and system-based executions~\cite{caffe}. However, none of the prominent CNN 
frameworks provide support for FPGA implementations. As a result, to implement a CNN on the FPGA, the designer has to manually design 
the implementation for each model, as well as test for correctness and optimize for performance, essentially rebuilding from scratch, 
rather than taking advantage of existing work.

CNNs are very computationally intensive with most of the computation in the convolution layers. 
The convolution layers require a large number of multiply-accumulate operations. 
The large computational complexity motivates efforts to reduce the number of required operations.
One approach is to use the FFT for convolution because in the frequency domain, the convolution becomes multiplication 
of each transformed input with the corresponding transformed filter coefficients, resulting in a compute reduction and speedup 
~\cite{fftconv}. An alternative approach uses the Winograd minimal filtering algorithm to take advantage of the overlapping 
computations between adjacent convolution windows~\cite{winograd1980arithmetic, nervana_wino}. In this work we implement and optimize the Winograd algorithm on an FPGA within
the Caffe framework~\cite{caffe}.

This paper makes the following contributions:

\begin{itemize}
\item We present an adaptation of the Caffe CNN framework with support for the Xilinx FPGA SDAccel environment. This
adaptation allows us to launch CNN classification on CPU-FPGA-based systems. 
\item We describe a modification to the Winograd convolution algorithm to further reduce DSP utilization for FPGA-based implementations. 
\item We implement the Winograd convolution algorithm targeting any $3\times3$ convolution layer with unity stride and benchmark it across several popular CNNs. Results show that the architecture achieves approximately 50 GFLOPS
across the $3\times3$ convolution layers of the benchmark suite, while using 83.2\% of the available SDAccel resources in a Xilinx Virtex 7 XC7VX690T-2. 
\item Finally, the software and hardware implementation details have been made open-source and can be found at the following link: \href{https://github.com/dicecco1/fpga_caffe}{https://github.com/dicecco1/fpga\_caffe}.
\end{itemize}

This work is organized as follows: Section~\ref{section:background} provides background information on CNNs and the Xilinx SDAccel OpenCL framework. Section~\ref{section:winograd_top} discusses the Winograd convolution algorithm and FPGA implementation. Section~\ref{section:fpga_caffe} details the features included in the FPGA Caffe framework and Section~\ref{section:results} shows the area utilization and performance results of the Winograd convolution engine within the FPGA Caffe framework. Section~\ref{section:rwork} reviews related work and compares this work to other recent FPGA implementations. Finally, Section~\ref{section:future_work} discusses future work related to the FPGA Caffe framework, and Section~\ref{section:conclusion} concludes the paper.

\section{Background}\label{section:background}

The following subsections detail the necessary background information regarding CNNs and the Xilinx SDAccel OpenCL development environment. 

\subsection{Convolutional Neural Networks}\label{section:CNNs}

CNNs are a popular type of supervised machine learning algorithm. Similar to other machine learning algorithms,
CNNs can be trained using back propagation to learn complex representations useful for many applications. CNNs are commonly used for
performing object recognition in pixel-based input. A popular CNN model such as AlexNet~\cite{alexnet} can be used to classify up to 1000
different objects in images with high accuracy.

These networks have two different modes of operation: training and inference. In the case of object recognition,
training involves feeding a large number of human-annotated images into the network. These images are used by the CNN to repeatedly
update the model's weights and biases such that it can learn how to recognize the objects in the human-annotated images.
Classification uses the trained CNN model and presents it images that the model has never seen to attempt to predict what the object in 
the image is. 

\subsection{CNN Layers}\label{section:cnnlayers}

AlexNet was one of the first successful ImageNet submissions employing CNNs and it consists of several layers: Convolution, Max Pooling, Fully Connected (FC), Rectified Linear Unit (ReLU), and Local Response Normalization (LRN)~\cite{alexnet}. In many of the top performing CNNs, Convolution, Pooling, Fully Connected, and ReLU layers are typically used, while LRN and other layers are sometimes used depending on the model~\cite{alexnet, VGG, OVERFEAT, GOOGLENET}. The general structure of a CNN usually consists of stacks of convolution layers with ReLU activations followed by a pooling layer.  In recognition applications, fully connected layers are used towards the output to reduce spatially organized information into a decision.  Convolution layers represent the majority of the computation for a CNN, with extreme cases requiring upwards of 30 GFLOPs of computation~\cite{VGG}. The ReLU layer is an activation function used to introduce a non-linearity into the network and can be described by $y=\max(0, x)$, which is applied to every data point of the input. Pooling is a simple reduction operation, such as max or average, applied to local regions of the input using a sliding window approach. FC layers consist of dense connections between neurons and usually contain the majority of the weights in the network. The computation of FC layers corresponds to a matrix multiplication followed by the addition of an optional bias parameter to each output. 

\subsection{Parallelism Strategies}\label{section:gpu_parallelism}

Given the large computational requirements and inherent parallelism of neural networks, these architectures are ideal for hardware accelerators.  
Popular parallelism strategies can be reduced to three main categories.

\noindent {\bf Data Parallelism} -- splitting the data across different execution threads, but using the same model. For the pixel-based
input (e.g.~images) natural to CNNs, data parallelism is inherent given the independence of individual and local groups of pixels.  
Fine-grained data parallelism can be applied using operations applied concurrently to all pixels, while coarse-grained data parallelism 
can be applied by processing ``minibatches'' of hundreds or thousands of input images during training.  

\noindent {\bf Model parallelism} -- splitting the model across different execution threads, but using the same data.  This strategy offers
several advantages, such as being able to accommodate large neural network sizes by splitting the weights across hardware accelerators, and
employing a type of efficient model averaging during training.  

\noindent {\bf Pipeline Parallelism} -- operating different dependent steps of computation concurrently on different threads, so that output
from one step is streamed as input to the next, while execution of steps is overlapping.  The feed-forward computation of CNNs are well suited 
for pipeline parallelism, so hardware that can exploit deep pipeline parallelism (e.g.~FPGAs) can offer an advantage. 

\subsection{SDAccel OpenCL FPGA Programming Model}\label{section:sdaccel}

\begin{figure}
\centering
\includegraphics[width=1\linewidth]{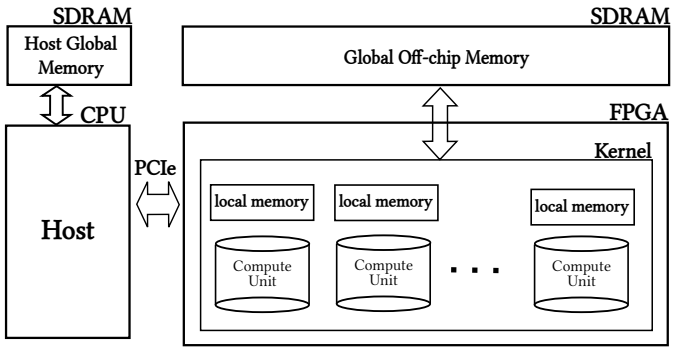}
\caption{SDAccel Platform}
\label{fig:platform}
\end{figure}

Using the SDAccel OpenCL environment to perform computations on an FPGA involves both host and kernel code. The host
code is used for programming the FPGA, passing data between the host's memory and the FPGA's global memory, and launching
the kernel on the FPGA. The FPGA is segmented into two regions, the programmable region and the static region. 
The static region is programmed upon power-up and it contains the interfaces to global memory and PCIe.
The programmable region contains the kernel, the computation to be accelerated.
The kernel code is synthesized into hardware and configured into the programmable region of the FPGA. 
The synthesized kernel can contain one or more compute units (CUs), where a CU
corresponds to the hardware unit responsible for the required computation. One approach to increasing parallelism in the
host code is to instantiate multiple CUs as shown in Fig.~\ref{fig:platform}, with each CU handling an
equally sized portion of the problem~\cite{sdaccel}. The portion of the problem handled by a single CU is referred
to as the local work group size, while the size of the overall task to be completed is referred to as the global work group
size. Each CU has its own local memory that is only accessible to the CU while all CUs share
the global off-chip memory of the platform. 

We have integrated SDaccel into the Caffe framework and used it to develop a convolution kernel using the Winograd transform.

\section{Winograd Convolution}\label{section:winograd_top}

Winograd convolution exploits the Winograd minimal filtering algorithm~\cite{winograd1980arithmetic}.
This approach has been shown to reduce the amount of required floating point operations~\cite{nervana_wino}. 
The sections below provide an overview for the Winograd convolution algorithm 
and discusses the implementation details of the FPGA-based Winograd convolution engine. 

\subsection{Winograd Convolution Algorithm}\label{section:winograd}

The Winograd convolution algorithm output is referred to as $F(m \times m, r \times r)$. 
In this expression $m \times m$ refers to the output tile size for a given input tile, meaning that
$m \times m$ output values are produced for every instance of $F(m \times m, r \times r)$. The filter
size in this case is $r \times r$. For a given filter size, many different values of $m$ can be chosen, 
which changes the computational complexity of $F(m \times m, r \times r)$, but does not impact the overall result. In this work, we implement a 
$F(2 \times 2, 3 \times 3)$ Winograd algorithm. In the case of $3\times3$ convolutions, $F(2 \times 2, 3 \times 3)$ has been shown to provide significant performance gains for GPU implementations in~\cite{nervana_wino}, though larger tile sizes may produce additional gains. This work targets $F(2 \times 2, 3 \times 3)$, mainly due to its simplicity of implementation however future work may explore other tile sizes as well. The equations for Winograd convolution are shown in Equations~\ref{eq:winograd_u} to~\ref{eq:winograd}. Equation~\ref{eq:winograd_u} shows how the filter transformation is calculated.

\begin{equation}\label{eq:winograd_u}
U = GgG^{T}
\end{equation}

\noindent Where\\
\begin{tabular}{rl}
$U$& is a $(m + r - 1) \times (m + r - 1)$ transformed filter;\\
$g$& is an $r\times r$ filter;\\
$G$& is an $(m + r - 1) \times r$ transform matrix, defined by \\  
 & the Winograd algorithm. \\
 \\
\end{tabular}

\noindent The filter values ($g$) are known at compile time and remain constant during run-time. 
Therefore, to save resources during run-time, 
the Winograd transformation for the filter values, shown in Equation~\ref{eq:winograd_u},
can be executed at compile time, on the CPU.
This approach saves FPGA resources.
However, pre-computing filter values increases the memory storage requirement. 
For direct convolution, the $3\times3$ filters require $C\times 3 \times 3$ storage,
where $C$ is the number of input channels.
For Winograd-based convolution, after the transformation, 
the $3\times3$ filter is transformed into a $4\times4$ matrix, 
requiring  $C\times4\times4$.
Therefore the storage requirement is increased by 33\%. 

Equation~\ref{eq:winograd_v} shows how the 
input transformation is calculated.

\begin{equation}\label{eq:winograd_v}
V = B^{T}dB
\end{equation}

\noindent Where\\
\begin{tabular}{rl}
$V$& is an $(m + r - 1) \times (m + r - 1)$ transformed data \\
 & tile;\\
$d$&is an $(m + r - 1) \times (m + r - 1)$ input tile;\\
$B$&is an $(m + r - 1) \times (m + r - 1)$ transform matrix, \\
 & defined by the Winograd algorithm.\\
\\
\end{tabular}

\begin{figure}[ht]
    \centering
    \includegraphics[width=0.6\linewidth]{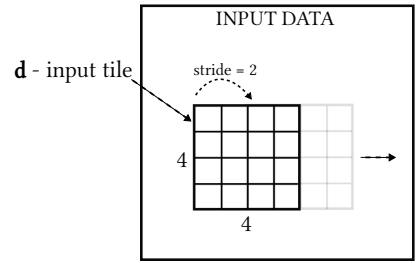}
    \caption{Winograd Input Tile Stencil}
    \label{fig:wino_full}
\end{figure}

\noindent For $F(2 \times 2, 3 \times 3)$ input tile ($d$) is $4\times4$ and is generated by a sliding window across the 2-D input feature data.
Shown in Fig.~\ref{fig:wino_full}, the $d$ window slides horizontally across the input data, 
with a stride of two. After the Winograd transformation, the $4\times4$ $V$ tiles are stored back into memory.
Because the input tiles contain overlaps of the input data, four times more memory storage is required. 

Equation~\ref{eq:winograd} shows how the pre-computed transformed filter $U$ 
and the run-time transformed input data $V$ are used to calculate the final output, 
a $2\times2$ tile $Y$. Each Y tile corresponds to a $2 \times 2$ non-overlapping subsection 
of the overall convolution output.

\begin{equation}\label{eq:winograd}
Y = F(m \times m, r \times r) = A^{T}[U \odot V]A
\end{equation}

\noindent Where\\
\begin{tabular}{rl}
$Y$&is an $m \times m$ output tile;\\
$\odot$& is an element wise multiplication;\\
$A$&is an $(m + r - 1) \times m$ transform matrix, defined by \\
 & the Winograd algorithm. \\
 \\
\end{tabular}

\subsection{FPGA Winograd Convolution}\label{section:fpga_wino}

To validate the FPGA Caffe framework, efforts were primarily focused on implementing the Winograd algorithm for convolution ($F(2 \times 2, 3 \times 3))$ discussed in Section~\ref{section:winograd}, though other less optimized implementations of the layers discussed in Section~\ref{section:cnnlayers} have been created as well. The architecture created for the Winograd algorithm on the FPGA can be separated into three stages of operation: input, compute, and output. 

{\bf Input Stage} -- used to move the input frames from off-chip SDRAM memory to the on-chip BRAMs. A portion of the input frame is burst read into a temporary buffer through an AXI request and moved into tiles of four rows by two columns stored in BRAM. The tiles in this case deviate from the algorithm discussed in Section~\ref{section:winograd} as data is only replicated in overlapping rows of tiles rather than both overlapping rows and columns to reduce the memory overhead of storing input tiles. This tiling method is similar to the method in Fig.~\ref{fig:wino_full}, however while the stride stays the same, the tile width is two resulting in no overlaps. To facilitate the compute stage and output stages, the number of tiles per row is set to be a multiple of eight, with padding added as required. After tiling is completed, a portion of the Winograd input transformation shown in Equation~\ref{eq:wino_col} is applied to every column of the $4\times2$ tiles.

\begin{equation}\label{eq:wino_col}
\begin{split}
O[0,j] &= I[0,j] - I[2,j]\\
O[1,j] &= I[1,j] + I[2,j]\\
O[2,j] &= I[2,j] - I[1,j]\\
O[3,j] &= I[1,j] - I[3,j]\\
\end{split}
\end{equation}

\noindent Where\\ 
\begin{tabular}{rl}
$I[n, m]$& is the input of the partial transform at $(n, m)$;\\
$O[n, m]$& is the output of the partial transform at $(n, m)$.\\
\\
\end{tabular}

This partial transformation is replicated eight times in the input stage such that eight columns of the input tiles can be processed per cycle to reduce the overhead that this preprocessing causes. 

\begin{figure}[ht]
    \centering
    \includegraphics[width=1\linewidth]{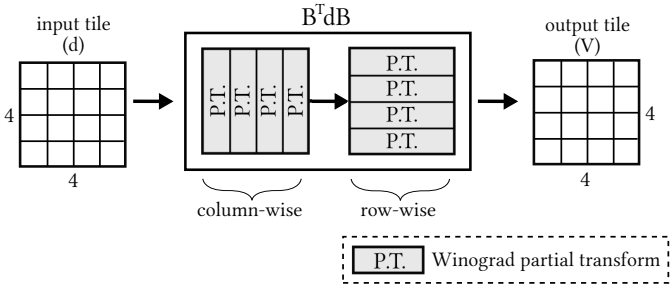}
    \caption{Winograd Forward Transform}
    \label{fig:wino_intro}
\end{figure}

The full input transformation in Equation~\ref{eq:winograd_v} requires a total of eight instances of Equation~\ref{eq:wino_col} per $4 \times 4$ tile, with four instances being applied to the columns, and four instances being applied to the rows of the tile (indices of Equation~\ref{eq:wino_col} are swapped for row calculations) as shown in Fig.~\ref{fig:wino_intro}. This results in 32 floating point additions and 64 DSPs per input tile transformation instantiation. Exploiting the fact that each input tile overlaps every two rows and every two columns with its neighbor, savings can be achieved
by first precomputing either all of the column wise instances of Equation~\ref{eq:wino_col} or all of the row wise instances rather than
computing the full transform for each tile. This reduces the number of partial transforms required for either the column wise or row wise
instances from four to two per tile, which reduces the number of floating point additions to 24 from 32 per input tile transformation (with potentially one additional set of column transformations for the edge of each row).

{\bf Compute Stage} -- where the bulk of the processing time is spent in the architecture. Following the input stage, each 4x2 tile and its neighbor are fed into a pipelined processing element that handles all subsequent computations required by the algorithm. The processing element completes the remaining set of partial transforms required after the input stage, performs the element wise multiplication between the input and the weights, computes the output transform, and accumulates the result within an output buffer. This process is repeated with different weights per output feature map until all of the output feature maps have been computed. Per output feature map this requires $C\times P\times Q + D$ clock cycles, where C is the number of input channels, P is the number of output tiles per column, Q is the number of output tiles per row, and D is the number of cycles required to fill the pipeline. To reduce the cycles required per output feature map, the processing element is replicated four times, allowing this stage to effectively be completed four times faster (assuming that D is small).  

{\bf Output Stage} -- handles transferring the output frame back into the off-chip DRR memory. First the results are gathered from the partial result buffers of the compute stage to an output buffer. Then the output is burst written to the DDR through AXI. 

To further improve the performance of the engine, it has been replicated such that there are two CUs rather than one. Each CU handles a 
separate input image to exploit coarse-grained data parallelism. The performance of the engine is directly proportional to how many CUs can be replicated, with the execution time being dictated by the equation $T = (C \times N)/ (F \times \#CUs)$, where F is the operating frequency, C is the number of clock cycles, N is the number of images, and T is the total latency. 

\section{FPGA Caffe Framework}\label{section:fpga_caffe}

The Caffe framework~\cite{caffe} is used to describe CNNs based on predefined layer implementations with CPU and GPU backends. 
In this section we describe our approach to augmenting the Caffe framework to enable CNN classification using FPGAs.
The discussion below will detail how memory transfers between the device and host are handled, how FPGA test
benches may be used within the framework, and several FPGA specific layer implementations. The layers include a custom layer for
reprogramming the FPGA and pipeline layers for fused layer implementations. 

\subsection{Caffe Model Description}

The infrastructure in Caffe allows for simple description of common layers used in CNNs and provides several
implementations of existing high performance CNNs as well. Each layer in Caffe corresponds to a set of computations
required by a given CNN model, allowing for modular CNN implementations. Caffe also allows for networks 
to be defined without modification of source code by providing model definitions through the Protocol Buffer Language
\cite{caffe}. This allows for networks to be constructed through a file describing which layers are
required and their respective ordering. 

The model description format in Caffe has been augmented in this work to support additional features described in 
the sections below. Namely, in the FPGA Caffe framework a program layer can be specified in any position of the 
network to force the FPGA to be reprogrammed, pipelined layers can be specified if a fused layer is 
required, and the Winograd convolution engine discussed in Section~\ref{section:fpga_wino} can be specified when
needed. 

\subsection{OpenCL Brew}\label{section:oclbrew}

In Caffe a Brew is referred to as a mode of operation that determines the target architecture on which CNN classification
or training is executed. The original Brews are CPU or GPU, with the CPU Brew containing the C++ infrastructure required to define layers
using a CPU, and the GPU Brew providing similar features but for NVIDIA GPUs using CUDA and cuDNN~\cite{caffe, cudnn}. For each Brew,
Caffe contains test cases available for every layer, allowing for fast determination of functional correctness and
benchmarking. 

\begin{figure}
\centering
\includegraphics[width=0.7\linewidth]{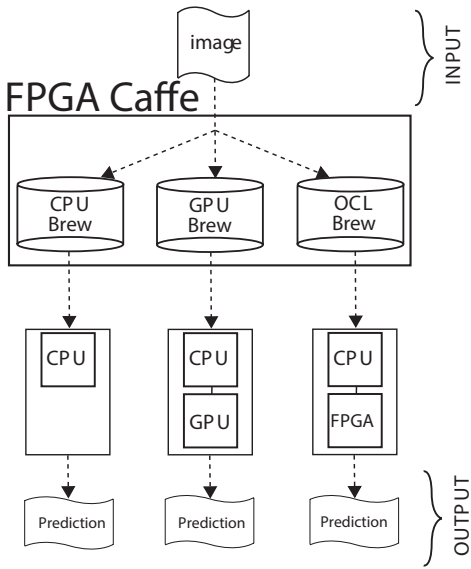}
\caption{High-Level View of the Brew Options in Caffe}
\label{fig:brews}
\end{figure}

This work extends the baseline Caffe framework to include the OCL (OpenCL) Brew, which provides support for Xilinx FPGA-based CNNs and could
easily be adapted to target Altera's OpenCL programming environment as well. The user can choose between the different Brews by
building the framework using the corresponding Makefile flags and changing the Brew to OCL. Fig.~\ref{fig:brews} shows an overview of the augmented system with
the OCL Brew, where inputs and outputs are the same as in the CPU and GPU Brews, but the underlying hardware of the system is comprised of the CPU for host code and the FPGA for
layer computations. 

To perform a forward pass (inference) using the OCL Brew, we added an API call: forward\_ocl(). The forward\_ocl() API
call is used as the forward operator on the condition that the Brew is OCL and the function is defined,
otherwise it defaults to the forward\_cpu() call as in the baseline Caffe implementation. 

\subsection{OpenCL Memory Management and Synchronization}\label{section:ocl_sync}

Data in Caffe is represented as a four-dimensional flattened array, with allocation, resizing, and synchronization
between CPU and GPU resources abstracted from its usage~\cite{caffe}. The memory management API in Caffe handles
synchronization between the host and GPU devices such that memory is only transferred back to the host when necessary. To
accomplish this, the state of the memory is stored as either HEAD\_AT\_GPU, HEAD\_AT\_CPU, or SYNCED which is verified upon
accessing the data. If the state of the data is HEAD\_AT\_GPU and the host requests the data, a data transfer from the
device to the host will be issued and the state will change to SYNCED. 

Support for memory synchronization between the host and the FPGA in the FPGA Caffe framework builds on the
memory synchronization features described above. To accomplish similar functionality, OpenCL
APIs are used with an additional object corresponding to the FPGA device memory object for each data structure. When data is
passed from the host to the FPGA, the state of the memory changes to HEAD\_AT\_OCL such that on subsequent accesses it will
either stay in the device memory or be transferred back to host memory. If the data is required by the host, a memory
transfer will be issued from the device to the host and the state of the memory will change to SYNCED. To access the FPGA
memory object, calls to either mutable\_ocl\_data() for modifying data (layer output data) or ocl\_data() for static data
(layer input data, such as weights), are required. These two functions were added to Caffe to handle both the creation and
synchronization of the device and host memory while maintaining transparency of memory manipulation as in the baseline
Caffe implementation.

\subsection{FPGA Testbenches}\label{section:testbench}

Testing a given layer in the FPGA framework can be accomplished in two different ways depending on the stage of development. 
Layers can be tested using individual test cases through the test framework provided in Caffe. Alternatively 
they can also be tested through the use of standalone host code by invoking only the host code required to launch 
the kernel. In either case, the layer can be tested using a hardware implementation or using software emulation based 
implementations created in the Xilinx SDAccel environment.

The baseline Caffe framework has a number of tests that are available for each layer of the system~\cite{caffe}. Each
test can be made into a test for the FPGA implementations by changing the Brew to OCL and modifying parameters to suit a
given layer. These tests allow for larger scale testing to verify that the layer has been integrated properly within the
Caffe framework. Aside from providing breadth to the test suite, this also allows for fast prototyping of layers through the
use of software emulated layers provided by the capabilities of SDAccel~\cite{sdaccel}. 

\subsection{Kernel Compilation}\label{section:compile}

Compiling kernels for CPUs and GPUs amounts to compiling programs into instructions that program the hardware, whereas compiling for FPGAs involves 
synthesizing full circuits. As a result, the overhead of compilation for FPGAs (hours) is much greater than that 
of CPUs and GPUs (milliseconds), and so runtime compilation of FPGA kernels is not possible.  We deal with this problem by employing
an offline compilation strategy, where deep learning practitioners can make use of precompiled binaries at run-time.  

\subsection{XCLProgram Layer}\label{section:program}

Though FPGA Caffe makes use of offline binary compilation, there is still significant overhead from programming the FPGA (100-300 ms)
compared to the CPU and GPU (0.001-0.005 ms).  This programming overhead conflates the measured execution time for a layer in 
the Caffe benchmarking functionality. We introduce a new layer, the XCLProgram layer, as a method for giving the user greater control over 
how the FPGA is programmed, as well as the ability to separately benchmark the execution time of each layer and programming overhead.
The XCLProgram layer as input receives a pointer to the FPGA binary file, as well as the kernel name. 

\subsection{Pipelined Layers}\label{section:pipeline}

In the GPU-based approach native to Caffe, modularity is enforced layer-wise, meaning before each layer is executed the GPU is programmed with the appropriate kernel
and memory is synchronized with the host.  In FPGA Caffe this becomes a bottleneck given the large overhead of programming the FPGA, and requiring such frequent 
memory synchronization with the host is much more expensive on FPGAs compared to high memory bandwidth GPUs.  Additionally,
this modularity used in Caffe limits parallelism strategies to within each kernel.  To address these issues, we introduce 
a new layer type in FPGA Caffe called pipeline layers.  Facilitated by XCLProgram layers, pipeline layers package multiple kernels into a
single binary, with kernel-kernel communication occurring through local memory structures on the FPGA (i.e.~FIFO).  Pipeline layers reduce
the number of times the FPGA is programmed, and eliminate the need to synchronize memory with the host between every layer.  Most importantly,
pipeline layers allow pipeline parallelism strategies across layers, increasing throughput by allowing multiple layers
to execute concurrently.  While the use of pipeline layers violates some modularity assumptions of Caffe, we argue that this is practical
given that combinations of layer groups are very predictable in practice (e.g.~convolution, ReLU, pooling).

\section{Results}\label{section:results}

This section describes our study of the Winograd Convolution Algorithm and the results gathered from using FPGA Caffe to implement the
$3\times3$ convolution layer using the Winograd Convolution Algorithm. The platform we use includes an Alpha-Data
ADM-PCIE-7V3 card with a Xilinx Virtex 7 XC7VX690T-2 running at 200 MHz and an Intel Xeon CPU
E5-2620 running at 2.0 GHz for the host application code. The Xilinx Virtex 7 is contained within
a server that has been virtualized to support virtual machines (VM) and is connected through PCIe. The VM in use 
has 8GB RAM and four cores. An Intel i7-4770k running at 3.5 GHz was used 
for CPU comparison and an nVidia Quadro K620 for GPU comparisons. 
The Xilinx SDAccel version number is 2015.1.3, CUDA version number is 7.5, cuDNN version 
number is 4.0~\cite{cudnn} and the CPU host code uses OpenBLAS~\cite{OPENBLAS} with eight threads enabled. The CPU, GPU, and FPGA
implementations all use single precision floating point as their data representation. 

\subsection{Winograd Resource Utilization}

The resource utilization post place and route is shown in Table~\ref{table:resources}. The highest utilization post place and route for both CUs
is the LUT utilization at 83.2\% of the SDAccel region's available LUTs. The utilization post place and route accounts for additional resources required to integrate into the SDAccel framework, which drives the significant LUT utilization in comparison to other resources on the device.

\begin{table}[ht]
\centering
\caption{Single Precision Winograd Convolution Engine Resource Utilization Post Place and Route}
\begin{tabular}{|c|c|c|c|c|c|l|} \hline
- & FF & LUT & DSP & \shortstack{BRAM \\ (18Kb)} \\ \hline
\shortstack{XC7VX690T \\ Total Resources (A)} & 866,400 & 433,200 & 3,600 & 2,940\\ \hline
\shortstack{SDAccel \\ Region (B) } & 551,040 & 275,520 & 2,376 & 1,940 \\ \hline
\shortstack{Winograd \\ Conv. Engine (C)} & 253,873 & 229,226 & 1,307 & 1,188\\ \hline
\shortstack{SDAccel \\ Utilization (C/B)} & 46.7\% & 83.2\% & 55\% & 61.2\% \\ \hline
\shortstack{Total \\ Device Utilization (C/A)} & 29.3\% & 52.9\% & 36.3\% & 40.4\% \\ \hline
\end{tabular}
\label{table:resources}
\end{table}

Given the utilization of the device resources, further instances of the CU could theoretically be added, though the 
limits on the available resources in the SDAccel region of the FPGA shown in Table~\ref{table:resources} makes it impossible to place
more than two CUs. This in turn limits the potential performance of the architecture within the SDAccel environment because of the overhead
of the static region and the lack of resources available in the reconfigurable region. 

To quantify the DSP savings from using the input stage column transformation discussed in Section~\ref{section:fpga_wino} we consider three
separate cases. The first case is that the full input transform is pre-computed for each four by four tile before sending the data to the
processing elements of the compute stage, with only one instance of the full input transformation in Equation~\ref{eq:winograd_v}. The second case is that the full input transformation in Equation~\ref{eq:winograd_v} is computed within each processing element of the compute stage using one full input transformation per processing element. Finally, the last case is the one discussed in Section~\ref{section:fpga_wino}, in which a column-wise partial transformation is computed for all input tile columns, using eight partial transformations to reduce computation overhead, with the remaining partial transformations computed within the processing elements using four partial transformations per processing element. Table~\ref{table:dsp_comparison} shows the resource utilization of the three cases post place and route. Between cases 2 and 3 there is a DSP savings of 61 units and a decrease in LUT usage. This is slightly less than what is anticipated in Section~\ref{section:fpga_wino}, though the difference can be attributed to fewer addressing calculations being required in case 2 due to the elimination of the partial transformations in the input stage. When comparing with case 1, the DSP utilization is approximately 20\% less than case 3, however the BRAM utilization has increased by 33\% due to the storing of replicated data in overlapping tiles. While the LUT usage and DSP utilization is better in case 1, the BRAM utilization makes it difficult to place and route more than 1 CU, as the BRAM utilization for two would require most of the available BRAM.

\begin{table}[ht]
\centering
\caption{Resource Utilization for Different Winograd Strategies With One CU}
\begin{tabular}{|c|c|c|c|c|c|} \hline
Layer & FF & LUT & DSP & \shortstack{BRAM\\ (18Kb)} \\ \hline
\shortstack{Case 1: Full \\ Pre-Transform} & 143,965 & 127,989 & 523 & 914 \\ \hline
\shortstack{Case 2: Full \\Transform in PE } & 166,905 & 153,887 & 715 & 688 \\ \hline
\shortstack{Case 3: Partial \\Transform in PE} & 158,266 & 145,892 & 654 & 688 \\ \hline\end{tabular}
\label{table:dsp_comparison}
\end{table}

\subsection{FPGA Caffe Benchmark Results}

To evaluate the Winograd convolution engine within the FPGA Caffe framework, a set of benchmark CNNs is required to view its performance across varying workloads (number of output feature maps and output sizes). The benchmark suite that we
use is adopted from the Soumith Chintala convnet-benchmarks~\cite{chintala}, which is composed of previous ImageNet winners including: AlexNet
\cite{alexnet}, VGG A~\cite{VGG}, Overfeat~\cite{OVERFEAT}, and GoogleNet~\cite{GOOGLENET}. Due to the RAM size of the virtual machine used for
the host code, the batch size for each benchmark is reduced by half to fit within the host VM. 

Table~\ref{table:timing} shows the performance of the system in comparison to both CPU and GPU implementations of the $3\times3$ convolution
layers of each CNN. To calculate the GFLOPS of the Winograd convolution engine, the number of floating-point operations is taken to be the same
as direct convolution, as is the case in~\cite{nervana_wino}, which is considered to be the effective GFLOPS. Comparing the geometric averages in Table \ref{table:timing}, the Winograd convolution engine performs approximately 2.1 times slower than the CPU implementation and 9.4 times slower
than the GPU implementation.

\begin{table*}[ht]
\centering
\caption{CPU, GPU, and FPGA 3$\times$3 Convolution Benchmark Results}
\begin{threeparttable}
\begin{tabular}{|c|c|c|c|c|c|c|c|c|c|c|l|} \hline
Network & \shortstack{CPU Run \\ Time (ms)} & \shortstack{GPU Run\\ Time (ms)} & \shortstack{FPGA Run \\Time (ms)} & \shortstack{CPU GFLOPS} & \shortstack{GPU GFLOPS} & \shortstack{FPGA GFLOPS}\\ \hline
AlexNet (64 Images) & 492 & 261.2 & 1,010 & 94 & 177.1 & 45.8 \\ \hline
VGG A (32 Images) & 4,310 & 745.14 \tnote{a} & 8,713 & 111.2 & 642.9 & 55\\ \hline
Overfeat (64 Images) & 2,030 & 387.1 & 4,781 & 139.2 & 730.2 & 59.1\\ \hline
GoogleNet (64 Images) & 1,506 & 209.66 \tnote{b} & 2,937 & 81.8 & 587.8  & 42\\ \hline
Geometric Average & 1,595.6 & 354.51 & 3,333.9 & 104.46 & 470.17 & 50\\ \hline \end{tabular}
\begin{tablenotes}
\item[a] GPU memory could only fit 8 images, so the model was run with 8 images and execution time was multiplied by 4 to get a performance estimate.
\item[b] GPU memory could only fit 32 images, so the model was run with 32 images and execution time was multiplied by 2 to get a performance estimate.
\end{tablenotes}
\end{threeparttable}
\label{table:timing}
\end{table*}

\section{Related Work}\label{section:rwork}

There exist several frameworks for implementing CNNs depending on the targeted platform and programming language,
with many of the popular frameworks discussed in~\cite{caffe}. These frameworks typically support a number of
programming languages including: C/C++, Matlab, Python, etc. With respect to Caffe, there is an independent effort underway
to add OpenCL support~\cite{oclcaffe}, though this support is meant primarily for exposing more GPUs to
Caffe rather than FPGAs. The work in~\cite{oclcaffe} provides additional functionality to support AMD
devices through OpenCL and provides some similar features to those explored in this work. However, these works are
differentiated in that the OpenCL implementations are abstracted away from being GPUs in this work through a separate Brew 
for OpenCL to allow for implementations across many different compute engines. Additionally, given the non-standard 
development model currently supported in FPGA OpenCL tools, frameworks that support standard OpenCL will not work with FPGAs without
significant framework modification. This is the case primarily because the FPGA OpenCL tools require offline compilation of kernels
and vendor-specific attributes to achieve suitable performance. The framework introduced in this paper allows for the support of FPGAs
through layer-specific implementations. These include the ability to specify when to program the FPGA, the addition of pipelined layers, and precompiled FPGA-specific layer implementations. 

Table~\ref{table:fpga_compare} shows the performance of this work compared to several recent FPGA works. The highest performing
implementations are Qiu~\cite{qiu} and Suda~\cite{suda}, which is enabled by their use of fixed point representations. The work of Zhang
\cite{zhang} is approximately 1.2-fold higher performance than this work while using the same data representation, though as expected our DSP
utilization is significantly lower while achieving comparable throughput. While the performance of this implementation is lower, it does not
require precision analysis prior to usage and it does not need to be resynthesized for new work loads as is the case in all of the prior works. 

\begin{table*}[ht]
\centering
\caption{Comparison Between Existing FPGA Works}
\begin{tabular}{|c|c|c|c|c|l|} \hline
Metric & Zhang~\cite{zhang} & Suda~\cite{suda} & Qiu~\cite{qiu} & This Work \\ \hline
Clock Frequency (MHz) & 100 & 120 & 150 & 200 \\ \hline
Precision & 32 bit float & 8-16-bit fixed & 16-bit fixed & 32 bit float \\ \hline
FPGA Version & Virtex 7 VX485T & Stratix-V GSD8  & Zynq XC7Z045 & Virtex 7 XC7VX690T-2 \\ \hline
DSP Utilization & 2,240 & (Not specified) & 780 & 1,307 \\ \hline
Host Connection & Microblaze, on chip host & PCIe & ARM Cortex-A9 Processor, on chip host  & PCIe \\ \hline
GFLOPS/GOPS & 61.62 & 136.5 & 187.8 & 50 \\ \hline
\end{tabular}
\label{table:fpga_compare}
\end{table*}

\section{Future Work}\label{section:future_work}

Future work related to this framework will contribute to both performance and usability.  First, completing the implementation of back
propagation will ensure FPGAs can be used for both classification and training, where hardware acceleration is especially important.
Given the modularity between the solver, network, and layer in Caffe, only the layers need to be modified to accommodate backward propagation. 
The structure of the network that defines the collection of layers, as well as the operation of the solver that calls the backward methods 
to generate gradients and perform a weight update, is already in place.

As well, experimenting with reduced-precision implementations of common layers is an important next
step. Recently, GPUs have started to support half precision based implementations of CNNs~\cite{cudnn, nervana_wino} 
and many of the FPGA works have been focused on reduced precision implementations
\cite{suda, qiu}. Half precision or fixed-point implementations both would offer significant area
gains for FPGA implementations as shown in~\cite{suda, qiu}, as a floating-point multiplication requires
three DSP units and additional LUTs and flip-flops  for current Xilinx FPGAs, while half precision or fixed-point
multiplications require only one DSP unit and significantly fewer LUTs and flip-flops. This would 
allow for the replication of both processing elements and CUs to improve performance. 

Finally, multi-FPGA-based parallelism strategies are crucially important in scaling up to accommodate larger data and model sizes. 
Current solutions involve using GPU clusters with Infiniband interconnects and MPI, which allow fast node to node data transfer 
and increase parallelism capabilities~\cite{coates2013deep}.  The use of FPGAs are attractive in this domain given the flexibility 
and high performance/watt, and FPGAs can benefit from much of the work being done investigating multi-GPU parallelism strategies.

\section{Conclusion}\label{section:conclusion}

In this work we presented a framework for implementing CNNs using FPGAs based on the Caffe CNN framework. The framework
allows for transparent support for individual FPGA implementations of layers for testing and verification. This framework
was validated by implementing the Winograd convolution layer and testing it across several CNNs. The results show that 
with 83.2\% of the available SDAccel resources we are able to achieve 50 GFLOPs across the $3 \times 3$ convolution layers of four 
different CNNs. While this does not improve upon current implementations in terms of performance, it demonstrates the capabilities of the
framework, which allows for further work that could lead to higher performance. 

\section*{Acknowledgment}

The authors would like to thank Xilinx, CMC and emSYSCAN, NSERC, and CFI for the funding and resources provided for this work.

\bibliographystyle{IEEEtran}

\bibliography{fpt}

\begin{thebibliography}{10}

\bibitem{zhang}
Chen Zhang, Peng Li, Guangyu Sun, Yijin Guan, Bingjun Xiao, and Jason Cong.
\newblock {Optimizing FPGA-based Accelerator Design for Deep Convolutional
  Neural Networks}.
\newblock In {\em Proceedings of the 2015 ACM/SIGDA International Symposium on
  Field-Programmable Gate Arrays}, FPGA '15, pages 161--170, New York, NY, USA,
  2015. ACM.

\bibitem{suda}
Naveen Suda, Vikas Chandra, Ganesh Dasika, Abinash Mohanty, Yufei Ma, Sarma
  Vrudhula, Jae-sun Seo, and Yu~Cao.
\newblock {Throughput-Optimized OpenCL-based FPGA Accelerator for Large-Scale
  Convolutional Neural Networks}.
\newblock In {\em Proceedings of the 2016 ACM/SIGDA International Symposium on
  Field-Programmable Gate Arrays}, FPGA '16, pages 16--25, New York, NY, USA,
  2016. ACM.

\bibitem{qiu}
Jiantao Qiu, Jie Wang, Song Yao, Kaiyuan Guo, Boxun Li, Erjin Zhou, Jincheng
  Yu, Tianqi Tang, Ningyi Xu, Sen Song, Yu~Wang, and Huazhong Yang.
\newblock {Going Deeper with Embedded FPGA Platform for Convolutional Neural
  Network}.
\newblock In {\em Proceedings of the 2016 ACM/SIGDA International Symposium on
  Field-Programmable Gate Arrays}, FPGA '16, pages 26--35, New York, NY, USA,
  2016. ACM.

\bibitem{caffe}
Yangqing Jia, Evan Shelhamer, Jeff Donahue, Sergey Karayev, Jonathan Long, Ross
  Girshick, Sergio Guadarrama, and Trevor Darrell.
\newblock {Caffe: Convolutional Architecture for Fast Feature Embedding}.
\newblock {\em arXiv preprint arXiv:1408.5093}, 2014.

\bibitem{fftconv}
Micha{\"{e}}l Mathieu, Mikael Henaff, and Yann LeCun.
\newblock {Fast Training of Convolutional Networks through FFTs}.
\newblock {\em CoRR}, abs/1312.5851, 2013.

\bibitem{winograd1980arithmetic}
S.~Winograd.
\newblock {\em {Arithmetic Complexity of Computations}}.
\newblock CBMS-NSF Regional Conference Series in Applied Mathematics. Society
  for Industrial and Applied Mathematics, 1980.

\bibitem{nervana_wino}
Andrew Lavin.
\newblock {Fast Algorithms for Convolutional Neural Networks}.
\newblock {\em CoRR}, abs/1509.09308, 2015.

\bibitem{alexnet}
Alex Krizhevsky, Ilya Sutskever, and Geoffrey~E. Hinton.
\newblock {ImageNet Classification with Deep Convolutional Neural Networks}.
\newblock In F.~Pereira, C.J.C. Burges, L.~Bottou, and K.Q. Weinberger,
  editors, {\em Advances in Neural Information Processing Systems 25}, pages
  1097--1105. Curran Associates, Inc., 2012.

\bibitem{VGG}
K.~Simonyan and A.~Zisserman.
\newblock {Very Deep Convolutional Networks for Large-Scale Image Recognition}.
\newblock {\em CoRR}, abs/1409.1556, 2014.

\bibitem{OVERFEAT}
Pierre Sermanet, David Eigen, Xiang Zhang, Micha{\"{e}}l Mathieu, Rob Fergus,
  and Yann LeCun.
\newblock {OverFeat: Integrated Recognition, Localization and Detection using
  Convolutional Networks}.
\newblock {\em CoRR}, abs/1312.6229, 2013.

\bibitem{GOOGLENET}
Christian Szegedy, Wei Liu, Yangqing Jia, Pierre Sermanet, Scott~E. Reed,
  Dragomir Anguelov, Dumitru Erhan, Vincent Vanhoucke, and Andrew Rabinovich.
\newblock {Going Deeper with Convolutions}.
\newblock {\em CoRR}, abs/1409.4842, 2014.

\bibitem{sdaccel}
Xilinx Inc.
\newblock {SDAccel Development Environment User Guide}, September 2015.

\bibitem{cudnn}
Sharan Chetlur, Cliff Woolley, Philippe Vandermersch, Jonathan Cohen, John
  Tran, Bryan Catanzaro, and Evan Shelhamer.
\newblock {cuDNN: Efficient Primitives for Deep Learning}.
\newblock {\em CoRR}, abs/1410.0759, 2014.

\bibitem{OPENBLAS}
Z.~Xianyi, W.~Qian, and Z.~Yunquan.
\newblock {Model-driven Level 3 BLAS Performance Optimization on Loongson 3A
  Processor}.
\newblock In {\em Parallel and Distributed Systems (ICPADS), 2012 IEEE 18th
  International Conference on}, pages 684--691, Dec 2012.

\bibitem{chintala}
Soumith Chintala.
\newblock Convnet-benchmarks.
\newblock [Online]. Available:
  \url{https://github.com/soumith/convnet-benchmarks}.

\bibitem{oclcaffe}
Fabian Tschopp.
\newblock {Efficient Convolutional Neural Networks for Pixelwise Classification
  on Heterogeneous Hardware Systems}.
\newblock {\em CoRR}, abs/1509.03371, 2015.

\bibitem{coates2013deep}
Adam Coates, Brody Huval, Tao Wang, David Wu, Bryan Catanzaro, and Ng~Andrew.
\newblock Deep learning with {COTS HPC} systems.
\newblock In {\em Proceedings of the 30th International Conference on Machine
  Learning ({ICML}-13)}, pages 1337--1345, 2013.

\end{thebibliography}

\end{document}